\DeclareUnicodeCharacter{2212}{-}
\documentclass[letterpaper, 10 pt, conference]{ieeeconf}  

\IEEEoverridecommandlockouts                              

\usepackage{xcolor}

\usepackage{graphicx} 
\usepackage{epsfig} 
\usepackage{times} 
\usepackage{amsmath} 

\usepackage{array}
\usepackage{ulem}
\usepackage{siunitx}
\usepackage{svg}

\usepackage{url}

\title{\LARGE \bf
 Scalable, Simulation-Guided Compliant Tactile Finger Design
}
\author{
    \authorblockN{Sandra Q. Liu, Yuxiang Ma, and Edward H. Adelson}
        \authorblockA{Massachusetts Institute of Technology\\
    {\tt\small sqliu@mit.edu, yxma@csail.mit.edu, adelson@csail.mit.edu}} 
}
\author{Yuxiang Ma$^{1,*}$, Arpit Agarwal$^{2,*}$, Sandra Q. Liu$^{1,*}$, Wenzhen Yuan$^{3}$, and Edward H. Adelson$^{1}$
\thanks{$^{1}$Yuxiang Ma, Sandra Q. Liu, and Edward H. Adelson are with the Massachusetts Institute of Technology
    {\tt\small sqliu@mit.edu, yxma@csail.mit.edu, adelson@csail.mit.edu}}
\thanks{$^{2}$Arpit Agarwal is with the Robotics Institute, Carnegie Mellon University
        {\tt\small \{arpita1\}@andrew.cmu.edu}}
\thanks{$^{3}$Wenzhen Yuan is with University of Illinois Urbana-Champaign
        {\tt\small \{yuanwz\}@illinois.edu}}
\thanks{$^{*}$Authors with equal contribution}
}    

\usepackage[colorinlistoftodos]{todonotes}

\newcommand{\gelsightfinray}{GelSight Fin Ray }
\begin{document}

\maketitle
\thispagestyle{empty}
\pagestyle{empty}
\begin{abstract}
Compliant grippers enable robots to work with humans in unstructured environments. In general, these grippers can improve with tactile sensing to estimate the state of objects around them to precisely manipulate objects. However, co-designing compliant structures with high-resolution tactile sensing is a challenging task. We propose a simulation framework for the end-to-end forward design of \gelsightfinray sensors~\cite{liu2022gelsight}. Our simulation framework consists of mechanical simulation using the finite element method (FEM) and optical simulation including physically based rendering (PBR). To simulate the fluorescent paint used in these GelSight Fin Rays, we propose an efficient method that can be directly integrated in PBR. Using the simulation framework, we investigate design choices available in the compliant grippers, namely gel pad shapes, illumination conditions, Fin Ray gripper sizes, and Fin Ray stiffness. This infrastructure enables faster design and prototype time frames of new Fin Ray sensors that have various sensing areas, ranging from \qty{48}{\mm} $\times$ \qty{18}{\mm} to \qty{70}{\mm} $\times$ \qty{35}{\mm}. Given the parameters we choose, we can thus optimize different Fin Ray designs and show their utility in grasping day-to-day objects. 
\end{abstract}

\section{Introduction}
Making robotic grippers more compliant and adaptive allows us to generalize grasping and robots' ability in unstructured environments. Robot grippers with integrated sensing have shown the ability to perform complex tasks~\cite{kuppuswamy2020soft,liu2022gelsight}. These grippers are integrated with high-resolution tactile sensing (640x480 texels) that allows in-hand rotation of mugs and manipulation of wine glass stems. 

However, designing compliant grippers with high-resolution tactile sensing is a nontrivial problem. Generally, high-resolution tactile sensors are made of rigid hardware components, which are difficult to incorporate into softer robotic bodies. 
Furthermore, manufacturing a single gripper can take at least a day predominantly due to elastomer curing times. This lengthy manufacturing process makes it difficult to perform rapid prototyping and generalize to different sizes, structures, and shapes of Fin Rays. In particular, we want to be able to make different tactile sensing Fin Rays and visualize their sensing regions before manufacturing one.

The ability to manufacture different Fin Rays is also useful from a utilization standpoint. Smaller Fin Rays are able to navigate through narrower spaces and singulate objects, while larger Fin Rays can apply more torque on an object and prevent larger objects such as mugs from rotating in their grasp. Having stiffer or softer Fin Rays can also help interact with bulkier or more fragile objects. Overall, we want to be able to streamline the design process through simulation of both mechanical and tactile structures and use this process to design and manufacture a variety of Fin Rays. 

As such, the main contributions of the work are three-fold:
\begin{itemize}
    \item We propose an end-to-end simulation pipeline, composed of FEM mechanical simulation and optical simulation with efficient fluorescent modeling. 
    \item We propose an efficient fluorescent material calibration setup and fit a low-dimensional fluorescent model for paints used in the Fin Ray gripper. 
    \item We show design variation along gel pad shape, illumination conditions, Fin Ray gripper sizes, and Fin Ray gripper stiffness.
\end{itemize}

\begin{figure}[ht!]
    \centering
    \includegraphics[width= .8\linewidth]{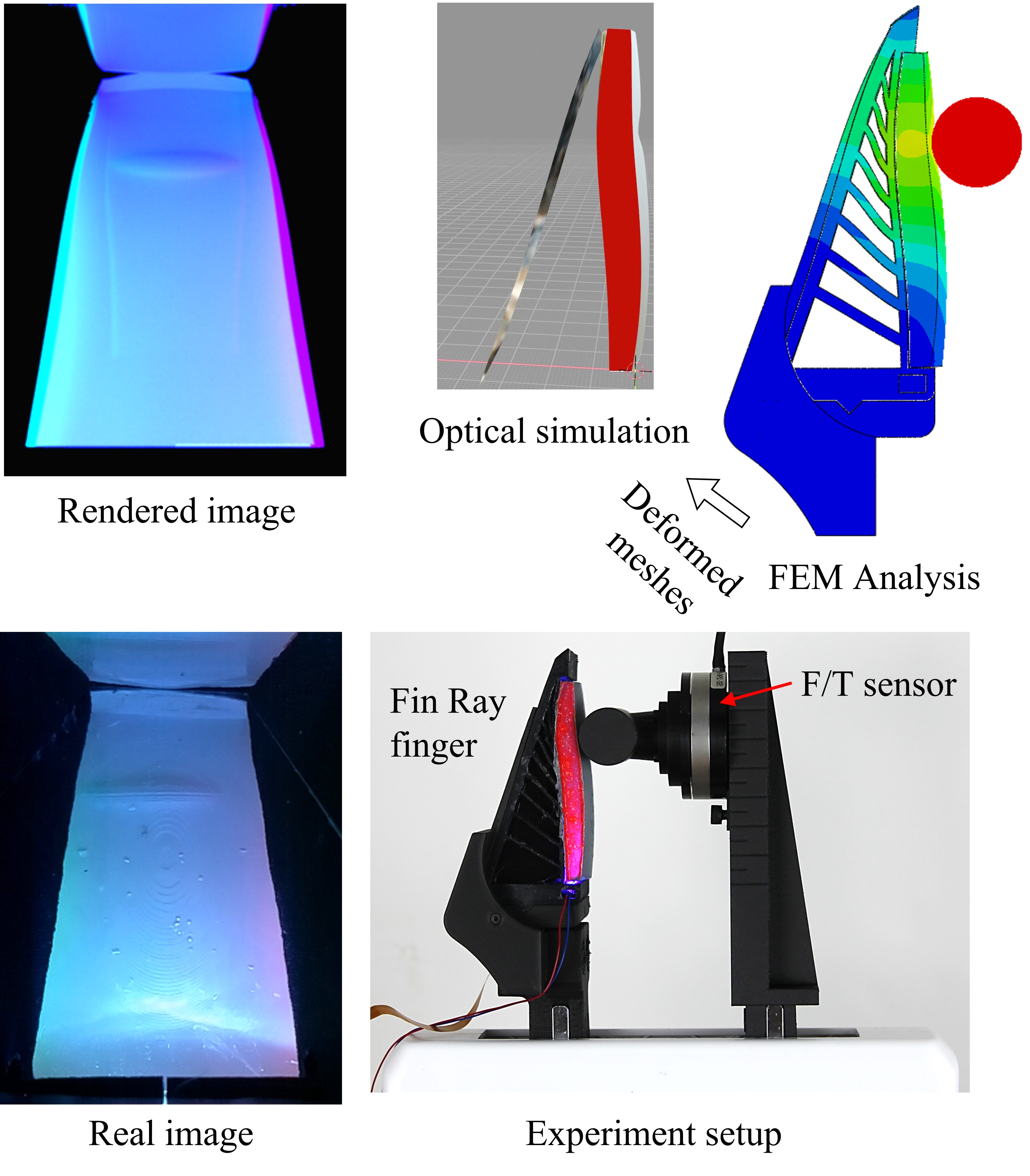}
    \caption{\textbf{Simulation-guided Fin Ray finger design}: New design of \gelsightfinray fingers can be made, based on the results of FEM mechanical simulation and optical simulation. The simulation provides a good prediction about the finger stiffness and tactile sensing performance.}
    \label{fig:teaser}
\end{figure}

\section{Related Work}

\subsection{Compliant Grippers with Sensing}

Many different soft robotic grippers and tactile sensors exist. Recently, there has been a surge of integrating tactile sensors with soft grippers, which generally rely on their mechanical structure or material properties to grasp a large variety of objects. Doing so enhances soft grippers so that they can perform a myriad of tasks using sensory feedback. These tactile sensors take on various form factors and generally provide lower-resolution tactile sensing.

One popular method of sensing is using strain sensors, as Georgopoulou et al. do in their 3D-printed soft robotic gripper. For ease of manufacturing, they integrate strain sensor elements using multi-material printing and are able to determine the size of the object the gripper is gasping \cite{strain3d}. Similarly, Chin et al. incorporate an elastic strain sensor on their handed shearing auxetic cylinders to provide proprioceptive information and add pressure sensors along the ``tactile'' surface for haptic sensing \cite{hsa_tactile}. 

Other methods of sensing involve using ``self-sensing,'' or sensing that measures the change of the soft robotic structure or actuator itself. An example of this concept is a shape-memory alloy compliant gripper, developed by Ganapathy et al., which is also able to sense its own deformation \cite{self-sensing}. Thuruthel et al. also designed a bistable soft gripper, which is able to rapidly and passively close when a force is applied on its body \cite{passive-sensing}. 

Another way of obtaining contact or tactile sensing is through microphones. Li et al. use microphones to pick up the resonant frequency of a hollowed-out channel inside a pneumatic sensor with contact sensing holes \cite{soft_mikes}. However, all of these tactile sensors are only able to determine low-resolution details such as if contact exists, or how much force is being pressed along an arbitrary part of the sensor. 

Recently, some works have begun to integrate camera-based tactile sensors into soft fingers, which allows for high-resolution contact sensing \cite{endoflex}. Nonetheless, incorporation of these sensors can be unwieldy in terms of electronic hardware and many soft fingers require additional hardware to actuate them. The Fin Ray is simple to actuate, can passively comply to many objects, and with the incorporation of fluorescent paints and flexible mirrors, the electronics hardware can be relatively simpler to integrate \cite{bbfinray}.  

\subsection{Fin Ray Gripper Design}
Fin Ray grippers are able to conform to complex shapes because of the Fin Ray structure, which consists of two long fins and horizontal ribs connected between them.  
The Fin Ray structure is inspired by the adaptive deformation of fish fins, known as the Fin Ray Effect \cite{alben2007mechanics}. Fin Ray grippers have shown great performance in grasping delicate and soft objects. They also have a higher payload than most soft robotic grippers, such as soft pneumatic grippers \cite{muller2020design}. 

Some researchers are devoted to designing Fin Ray fingers that are able to grasp a wide range of objects. Crooks et al. improved the Fin Ray's payload by combining soft and hard plastics in their Fin Ray finger \cite{crooks2016fin}. Shin et al. proposed a Fin Ray gripper design with an improved weight capacity and an expanded range of graspable geometries by attaching friction pads and enabling parallel and centric gripping modes \cite{shin2021universal}. Elgeneidy et al. investigated the passive stiffening of Fin Ray fingers due to layer jamming and presented a soft Fin Ray finger that can perform both delicate and high-force grasping. There is also research which optimized different Fin Ray structure parameters, including number, thicknesses, and angles of the ribs, to obtain higher grasping/wrapping quality \cite{deng2021learning, suder2021structural}. 

In our prior research on the GelSight Fin Ray finger, we first incorporated vision-based tactile sensing into the Fin Ray finger by hollowing out the Fin Ray skeleton \cite{liu2022gelsight}. The GelSight Fin Ray finger retains good compliance after the integration, but the hollowed out finger structure leads to inadvertent twisting motions. To improve the robustness of the finger, we designed the GelSight Baby, in which we placed the camera at the finger base and synthesized a silicone adhesive-based fluorescent paint to give the finger better compliance \cite{bbfinray}. 

In this work, we design Fin Ray fingers with different sizes and stiffness values based on simulation results. Fin Ray structures can be easily scaled up or down to grasp a variety of objects with varying forces and torques. Without changing materials, the stiffness of Fin Ray fingers can be tuned by the rib arrangement in the Fin Ray structure. Moreover, we are able to achieve various stiffness by changing the rib number and modifying the width and thickness of ribs. In our previous work \cite{bbfinray}, we have established FEM models to analyze the compliance of different Fin Ray designs. In this work, we combine FEM simulation with optical simulation to provide realistic previews of tactile images before building prototypes. However, defining a new FEM model takes several or more hours because the components of the \gelsightfinray finger have complex geometries and contact relationships. Therefore, we established a parametric modeling pipeline using the Python-based ABAQUS Scripting Interface to massively generate FEM models. 


\subsection{Tactile Sensor Simulation}
The most recent works on tactile simulation focus on finding an efficient approximation of soft body simulation by a penetration depth-based linear model \cite{wang2022tacto} or FEM-based \cite{narang2021interpreting}. In  Xu et al's paper \cite{xu2023efficient}, the authors built a differentiable tactile simulator for marker motion. However, all the above simulators are focused on contact dynamics simulation. In Si et al. \cite{si2022taxim}, authors developed contact dynamics based on a linear elasticity model and tracking marker motions. They also generate images based on a few datasets collected from a real sensor. However, none of the above work focuses on performing optical simulation of novel sensors. In Agarwal et al. \cite{agarwal2021simulation}, authors used physically based rendering for simulating GelSight \cite{yuan2017gelsight}. Their method is fully general and allows the simulation of an arbitrary sensor without collecting any sensor-specific data. Our work also uses physically based rendering with the addition of an approximate fluorescent rendering technique. The fluorescent rendering is essential for an accurate simulation of GelSight Fin Ray sensors.    


\section{Methods}

\subsection{Hardware}

\begin{figure}[ht!]
    \centering
    \includegraphics[width= 0.8\columnwidth]{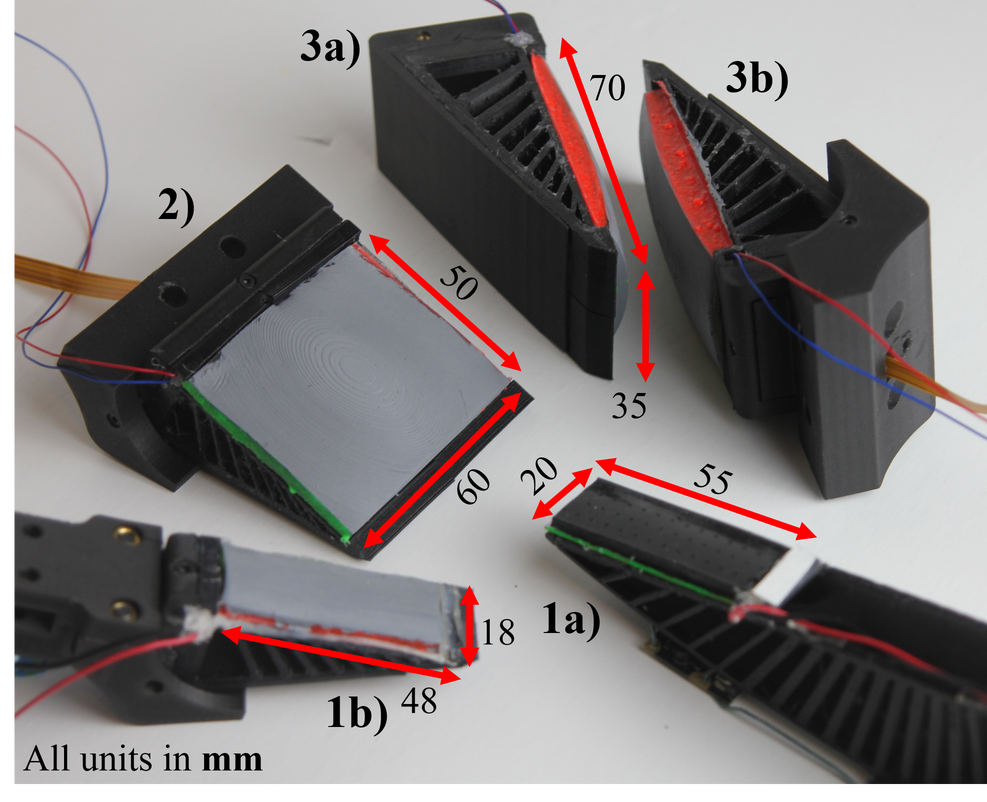}
    \caption{\textbf{The family of GelSight Fin Rays}:  1a) is our original one, and 1b) is the GelSight Baby Fin Ray. 2) represents a box like Fin Ray, which can be more useful for applying torques to long tool handles, while 3) represents two versions of a longer and medium-width Fin Ray, on which we performed simulations. 3a) represents a stiffer Fin Ray structure, while 3b) represents a softer structure with thinner ribs.}
    \label{fig:finrayfamily}
\end{figure}

We scale the GelSight Baby Fin Ray so that the length of the pad is \qty{70}{\mm} and the width of the pad is \qty{35}{\mm}, shown by Fig. \ref{fig:finrayfamily} 3). This \qty{35}{\mm} $\times$ \qty{70}{\mm} Fin Ray size is chosen so that the Fin Ray will be more useful for grasping and exerting torque on larger objects and will create more surface area for a stabler grasp. The thickness of the gel pad is also scaled accordingly with respect to the gel pad width. 

All other manufacturing processes were kept the same. We print the Fin Ray structure out of TPU 95A and include a semi-rigid Onyx backing (Markforged) to help mitigate twisting of the hollowed-out Fin Ray. Laser-cut cloth is used on the sides to prevent environmental lighting interference. A 120$^{\circ}$ field of view Raspberry Pi camera, which sits at the base of the structure, views a \qty{0.15}{\mm} PET-G flexible mirror, which in turn reflects the tactile sensing area. 

Because the tactile pad is thicker than the gel pads on the two prior GelSight Fin Rays, we choose to use a softer gel pad. Using a softer pad will allow us to still pick up on finer details in tactile resolution even if the silicone is thicker. The gel pads are made using a mixture of 1 part XP 565 catalyst (Silicones Inc.) to 15 parts XP 565 to 3 parts plasticizer (LC1550 Phenyl Trimethicone, Lotioncrafter). We utilize the same aluminum cornflake silicone mixture (\qty{35}{\mu\m} aluminum cornflake, Schlenk Offset FM/6500) for the tactile sensing surface which is airbrushed onto our molds before we add the silicone mixture. After curing, we paint the sides of the silicone gel pads with red or green fluorescent acrylic paint and silicone adhesive (A-564, Factor II) mixture. The Chanzon blue LEDs at the base of the Fin Ray illuminate the gel pad and cause the paint to fluoresce, giving us a tri-color sensing region. 

The silicone gel pads are created using various 3D-printed high-resolution molds. For our cylindrical and flat molds, we adhere a \qty{0.1}{\mm} mylar sheet to the bottom of the molds. Doing so allows us to have a smooth tactile sensing surface. We are unable to do the same for our ellipsoid molds, but attempted to put a thicker layer of aluminum silicone paint to counteract the visible striations formed by the 3D-printed surface. We also note that this effect could be mitigated by polishing or sanding the surface of the mold. 

To show the usefulness of the simulation on other GelSight Fin Ray sizes, we also use our GelSight Baby Fin Ray, a less stiff version of the \qty{35}{\mm} $\times$ \qty{70}{\mm} one, and another Fin Ray with a tactile sensing pad that is \qty{50}{\mm} in height and \qty{60}{\mm} in width. The Baby Fin Ray, as shown by Fig \ref{fig:finrayfamily} 1b), will be more useful in smaller manipulation applications, while the wider Fin Ray in Fig \ref{fig:finrayfamily} 2) will be able to apply more torque on different handheld tools. On the other hand, the less stiff Fin Ray in Fig \ref{fig:finrayfamily} 3b) will be more useful for grasping softer objects. 

\subsection{Optical Simulation}
Our optical simulation pipeline is based on physically based rendering (PBR), which simulates light transport inside the sensor with accurate models of light, refractive surface, fluorescent paint, and coating material. We built on tactile simulation techniques introduced in Agarwal et al.~\cite{agarwal2021simulation} based on PBR. One of the key differences in the GelSight Fin Ray is the use of fluorescent paint for illumination. Therefore, in this section, we introduce the fluorescent material model and simplified fluorescent simulation technique in PBR. In the first part, we will describe the simplified model for fluorescent paint and its calibration process. In the second part, we will describe the simplified simulation model of fluorescence lights in the PBR framework that allows us to simulate the full sensor. 

\begin{figure}
    \centering
    \includegraphics[width=\columnwidth]{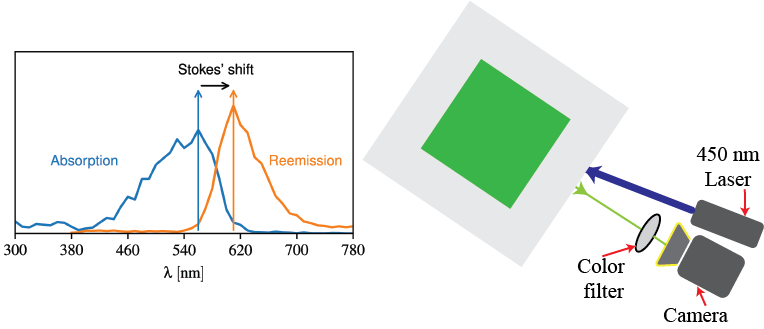}
    \caption{\textbf{Fluorescent model and  calibration setup}: (A) shows a canonical fluorescent material model~\cite{hua2023efficient}. It consists of absorption and reemission spectra whose peaks are separated by Stokes' shift. (B) shows the imaging setup we created to capture the reflectance at the excitation wavelength, $\lambda=450 \text{nm}$ for calibrating fluorescent paints used in GelSight Fin Ray.}
    \label{fig:fluo_capture}
\end{figure}

\paragraph{Fluorescent paint calibration} For accurate simulation, we need to calibrate the fluorescent paints used in the sensor. Our imaging setup consists of a CM3-U3-13Y3C-CS 1/2" Chameleon color camera, \qty{450}{\nm} Blue Alignment Laser Diode Module (Edmund Optics) and 8 color filters with central wavelengths -- \qty{405}{\nm}, \qty{450}{\nm}, \qty{500}{\nm}, \qty{532}{\nm}, \qty{560}{\nm}, \qty{600}{\nm}, \qty{630}{\nm}, and \qty{660}{\nm}. We calibrated two fluorescent paints (Liquitex BASICS Acrylic Paint Red Fluorescent ASIN B07F48YG5F and Liquitex BASICS Acrylic Paint Green Fluorescent ASIN B07F48WZWL) made in a flat sample. The calibration setup is shown in Figure \ref{fig:fluo_capture}. We assumed that the fluorescent paint is diffuse in nature -- for any incident direction, the amount of outgoing light radiance remains the same. 

\begin{figure}
    \centering
    \includegraphics[width=\columnwidth]{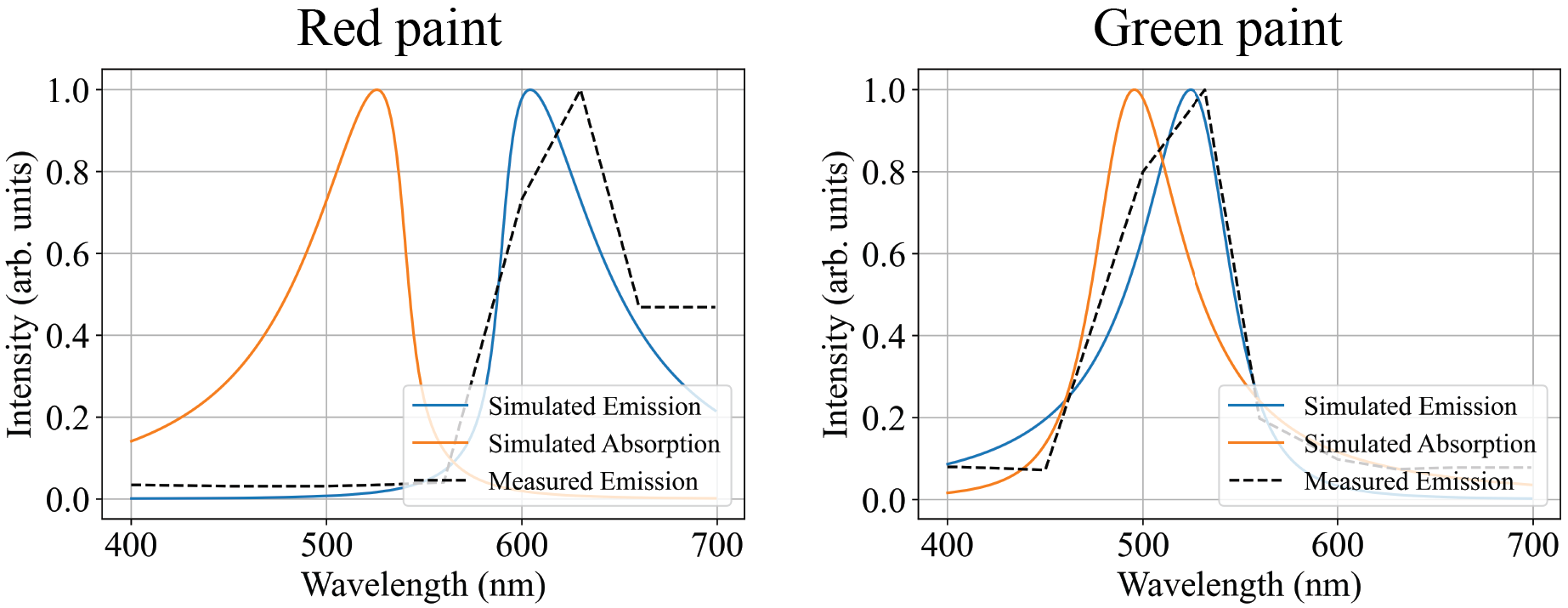}
    \caption{\textbf{Fluorescent paint fit}: This shows the comparison of measured emission spectra and simulated emission spectra using a 4D parametric model for red fluorescent paint (left) and green fluorescent paint (right).}
    \label{fig:fluo_fit}
\end{figure}

A fluorescent material could be characterized by absorption and emission spectra. The first defines which incident light wavelengths are absorbed and lead to reemissions. The second describes the amount of reemission across all incident wavelengths. The difference between the spectral positions of the band maxima of absorption and reemission is called a Stokes shift~\cite{mcnaught1997compendium}.  
According to Zheng et al. ~\cite{zheng2014spectra}, if the spectra is not very spiky, absorption and emission spectra can be modeled by a 4-parameter analytic distribution, a variant of skew Cauchy distribution. The spectra value at wavelength $\lambda$ is given by the function   
\begin{multline}
    \text{f}(\lambda|\lambda_0,\gamma, \omega, \text{h}) = \\
    \frac{\text{h}}{\left[\gamma^2+(\lambda-\lambda_0)^2\right]} \Biggl\{\frac{1}{\pi} \arctan \biggl[ \frac{\omega(\lambda-\lambda_0)}{\gamma} \biggr] + \frac{1}{2}\Biggl\}
\end{multline}
where $\lambda_0$ is the peak wavelength, $\text{h}$ is height parameter, $\gamma$ is width and $\omega$ is the skewness parameter.   
We manually fit the measured data and choose Stoke's shift for the paint to be \qty{100}{nm} and \qty{50}{nm} for red and green fluorescent respectively, based on reflectance data. For calibrating the non-fluorescent reflectance, we collected images in room light and matched them to the closest color in a traditional colorchecker. We found the non-fluorescent reflectance of the red and green fluorescent paint is very similar to colorchecker \textit{Red} and \textit{Green} colors respectively.

\begin{figure}
    \centering
    \includegraphics[width=0.3\columnwidth]{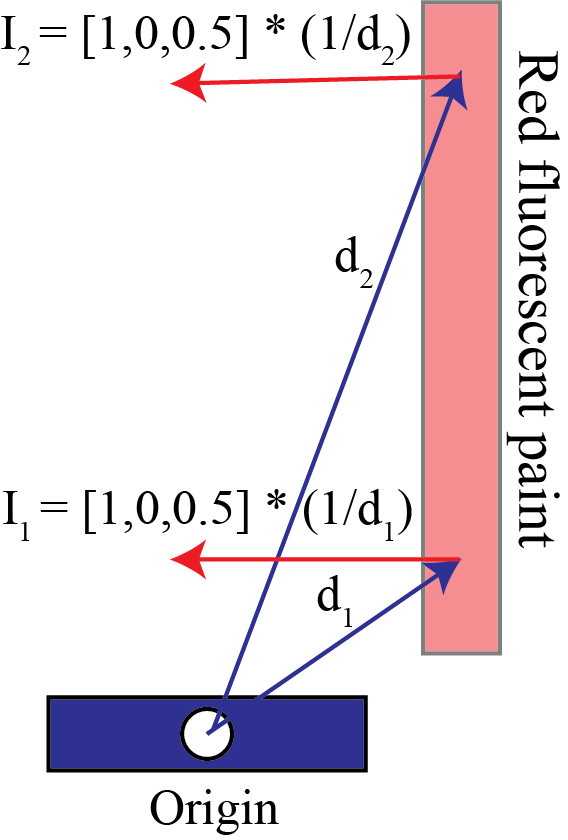}
    \caption{\textbf{Fluorescent rendering}: This image shows the visual of the efficient rendering of fluorescent paint lights using our parametric reflectance model for simulating GelSight Fin Ray sensor.}
    \label{fig:fluo_rendering}
\end{figure}

\paragraph{Fluorescent simulation model} We found the fluorescent effect leads to reflectance of around 2\% - 5\% in the desired wavelength. Additionally, it depends on the incident excitation wavelength. For a fast approximate model, we created a textured light source whose intensity is proportional to the distance from the center of the blue light source. The color of the LED is chosen based on our calibration model in the previous section. Pictorially, the fluorescent light source looks as shown in Figure \ref{fig:fluo_capture}.
Our components modeled in the optical simulation are shown in Figure \ref{fig:fluo_rendering}. Thereafter, we use GPU path-tracing to generate all the images. 

\subsection{Mechanical Simulation}

\begin{table*}[!htb]
    \centering
    \caption{Constitutive models and element types of all components }   \label{tab:material}
    \begin{tabular}{c|cccc}
    \hline
    Component & Material & Constitutive Model    & Parameters & Element Type \\ \hline
    Fin Ray skeletons & SainSmart TPU 95A          & Ogden model \cite{TPU_model}          &    \begin{tabular}{@{}c@{}}$\mu_1 = 6.279 \text{MPa}, \mu_2 = 1.639 \text{MPa}$\\ $ \alpha_1 = 1.6663, \alpha_2 = −7.136$ \cite{TPU_model} \end{tabular} &  C3D8R\\   
    Silicone pad & PDMS               & Neo-hookean model & $C_{10}=0.1333 \text{MPa}$  & C3D8R \\
    Base, backing & Onyx  & Linear elastic model  & $E=2.1 \text{GPa}, \nu = 0.38$ & C3D8R \\
    Flexible mirror  & PET-G   & Linear elastic model  & $E=2.8 \text{GPa}, \nu = 0.4$ & CSS8 \\
    Mylar sheet & Mylar               &   Linear elastic model  & $E=5 \text{GPa}, \nu = 0.38$ & CSS8 \\    
    Indenters & Onyx & Rigid & N/A & R3D4 \\ \hline
    \multicolumn{5}{l}{ }\\
    \multicolumn{5}{l}{Ogden model is a hyperelastic model with Helmholtz free energy defined as $\Psi = \sum^{N}_{p=1} \mu_p$/$\alpha_p*( \lambda^{\alpha_p}_1 + \lambda^{\alpha_p}_2 + \lambda^{\alpha_p}_3 )$ , where} \\
    \multicolumn{5}{l}{ $\mu_p$ and $\alpha_p$ are material constants, and $\lambda_i, i=1,2,3$ are principle stretches. $E$ and $v$ are Young's modulus and Poisson's ratio} \\
    \multicolumn{5}{l}{in linear elastic model.}
    \end{tabular} 
    \vspace{-5 mm}
\end{table*}

We use FEM to simulate the mechanical interaction between Fin Ray fingers and indenters. The deformation of Fin Ray fingers are used to evaluate the fingers' stiffness and provide deformed sensing pads for optical simulation. The models are created in Abaqus (Simulia, Dassault Systemes). 

\begin{figure}[ht!]
    \centering
    \includegraphics[width=0.95\linewidth]{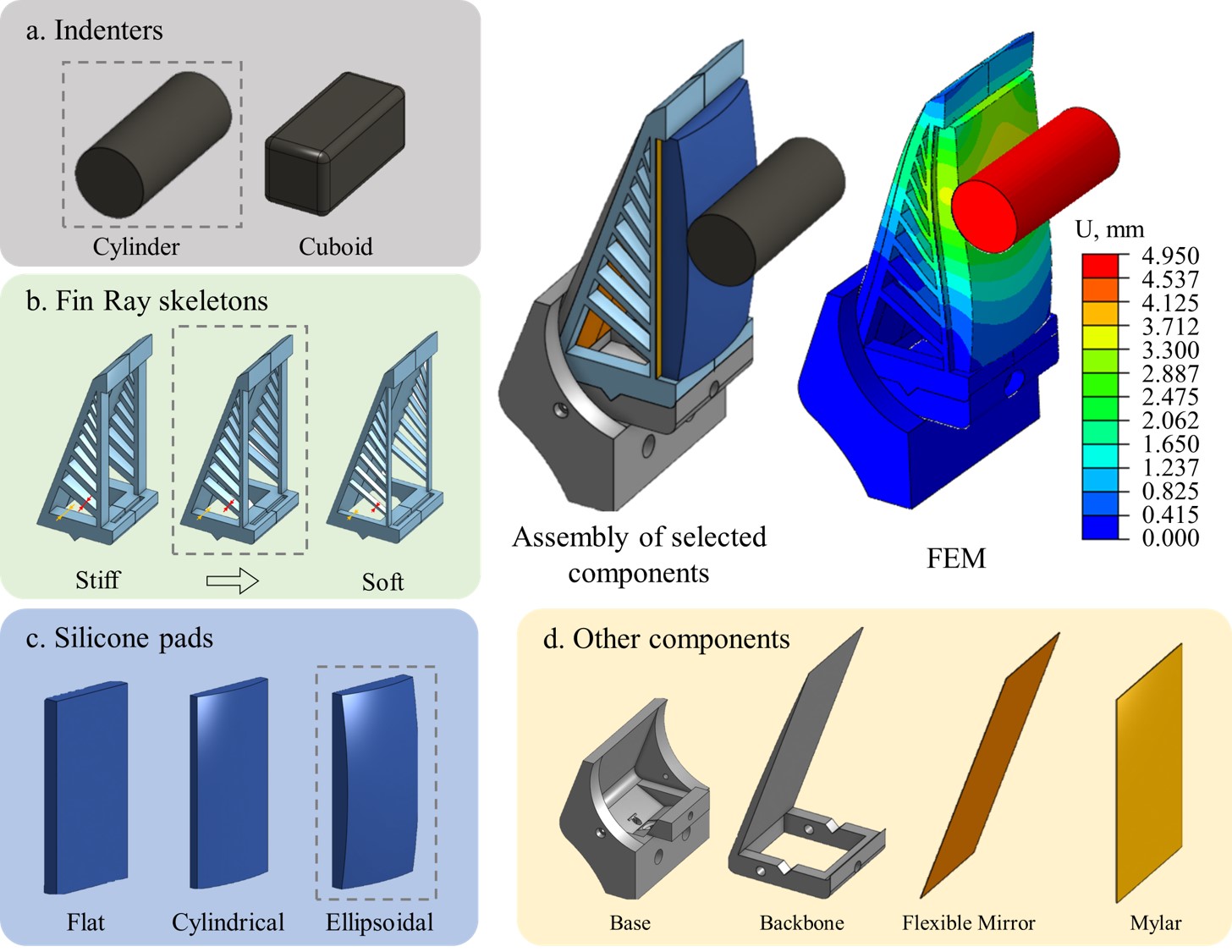}
    \caption{\textbf{Parametric modeling pipeline of FEM simulation}: New FEM models can be created by selecting different components from candidates. Fin Ray design with optimal illumination and desired stiffness can be found by simulating different combinations of Fin Ray skeletons and silicone pads. }
    \label{fig:FEM}
\end{figure}

\paragraph{Parametric modeling}
Abaqus Scripting Interface provides a way to define and modify all modules of an Abaqus model, such as parts, materials, interactions, loads, and meshes, known as parametric modeling. Using the interface, we are able to select and combine parts to create FEM models and analysis jobs automatically, as shown in Fig. \ref{fig:FEM}. 

In a Fin Ray finger assembly, the two most important components are Fin Ray skeletons and silicone pads. Different stiffness Fin Ray fingers are created by modifying the rib distribution as well as the thicknesses and widths of the skeletons, while better sensing illumination can be acquired by changing the shapes of the silicone gel pads. Tie constraints are used to assemble the selected Fin Ray skeleton and silicone pad with other structural components. After that, we apply displacement load to a cylindrical or cuboidal indenter to deform the Fin Ray finger. We chose these indenters as they provide a wide range of surface normals and are easy to analyze.   Surface contact is defined for all possible possible contact pairs, e.g. between the silicone pad and the indenter, between adjacent ribs of the skeleton. 

Table \ref{tab:material} summarizes material constitutive models and element types of different components in the Fin Ray design. Since Onyx, PET-G, and Mylar are about three orders of magnitude stiffer than TPU 95A and PDMS, small deformation assumption and linear elastic model are adopted for their corresponding components. We assume indenters are rigid to simplify their interaction with silicone pads, because soft-right contact is computationally less expensive than soft-soft contact. While the indenters are fabricated using Onyx, their minimal deformation does not result in noticeable displacements such as backbone deflection, thereby supporting the assumption of rigidity. TPU 95A and PDMS are soft nonlinear hyperelastic materials, which may experience large deformation during grasping. Ogden model is able to fit the strain-stress curves of commercial TPU filaments, and \cite{TPU_model} offers the Ogden model parameters of Sainsmart TPU 95A, while Neo-hookean model can effectively capture the non-linear elasticity of PDMS \cite{PDMS_model}. C3D8R solid elements are assigned to TPU, PDMS, and Onyx, CSS8 shell elements are assigned to the PET-G mirror sheet and Mylar sheet, and R3D4 rigid elements are assigned to the indenter. To capture more deformation details on the silicone pad, we mesh the silicone pad and the Mylar sheet with a small element size of \qty{0.4}{\mm}, while the other components are meshed with bigger element sizes between \qty{0.6}{\mm} to \qty{1.5}{\mm}.

The models are analyzed with a nonlinear static step, so that nonlinear elasticity and nonlinear geometric contacts can be well simulated. A finer mesh is used to verify that the simulation converges well. Deformed node coordinates of the flexible mirror, the Mylar sheet, and the silicone pad are exported to an Abaqus results file (*.fil), which is used to generate surface meshes for optical simulation. 

\paragraph{Mesh conversion}
Optical simulation requires triangular surface meshes of the silicone pad, the Mylar sheet, and the flexible mirror, but hexahedral volumetric meshes are applied to these components in FEM. We established a mesh conversion protocol to deal with this mesh mismatch. First, we use Abaqus2Matlab \cite{abaqus2matlab} to load node coordinates as well as element information from the exported Abaqus results file. After that, we find all boundary nodes by sorting node coordinates and then find all boundary elements that include these nodes. Then, we find all boundary quadrilateral faces of the hexahedral elements and divide these quadrilateral surfaces into two triangular faces. Finally, the nodes in each triangular face are resorted to keep the normal directions consistent. In this way, we create high-quality triangular meshes for deformed gel pads. 

\section{Experiments}
\subsection{Shapes of Silicone Gel Pad}

\begin{figure}[ht!]
    \centering
    \includegraphics[width=0.9\linewidth]{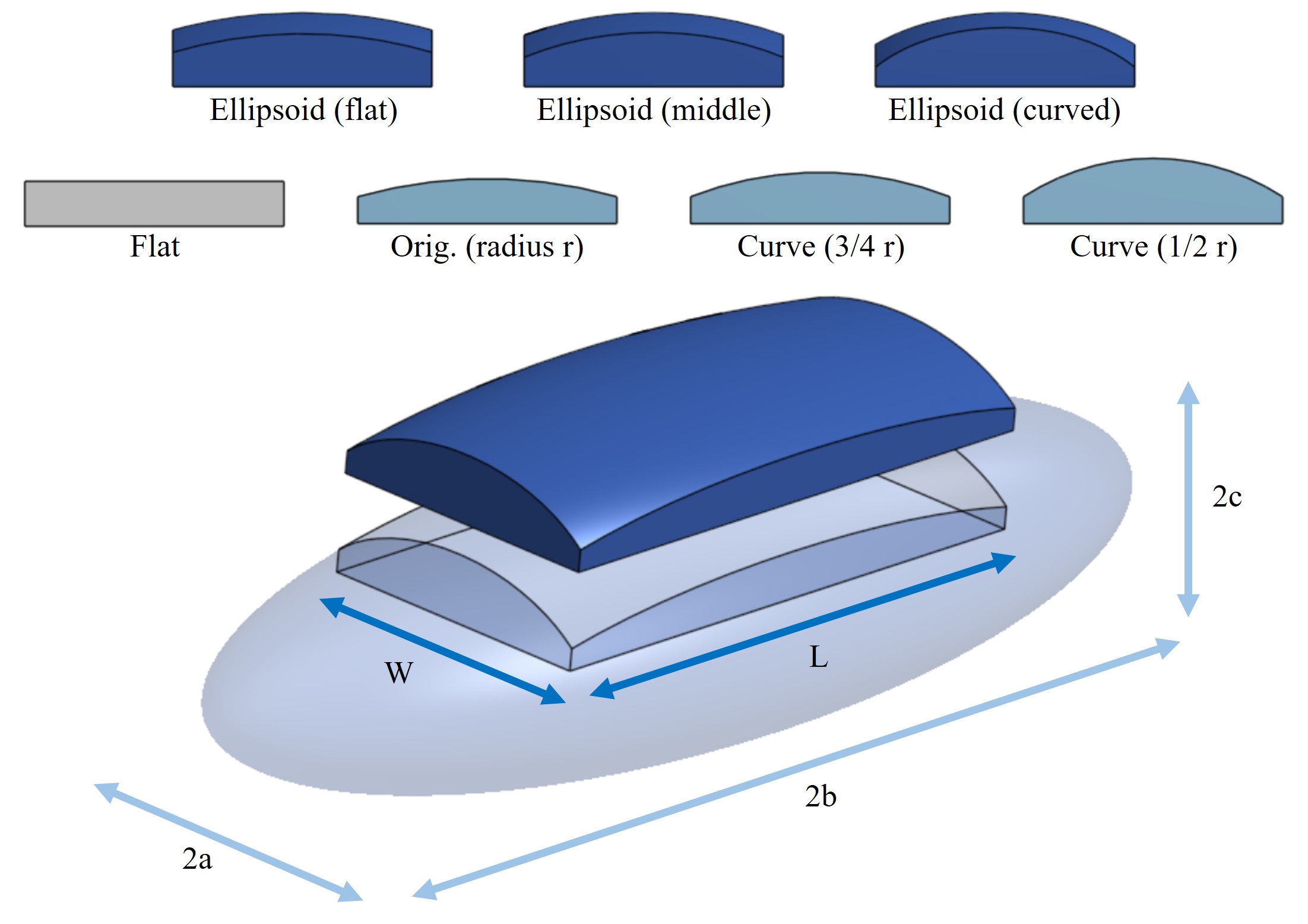}
    \caption{\textbf{Different gel pad profiles}: On top are the different gel pad profiles we simulated, which are color-coded by the type of shape they have. On the bottom is an example of how we create the shape of our ellipsoid gel pads with the corresponding ellipsoid and gel pad dimensions.}
    \label{fig:gelpads}
    \vspace{-4 mm}
\end{figure}

We consider three different types of silicone gel pad shapes: cylindrical, ellipsoid, and flat (Fig. \ref{fig:gelpads}). These shapes were considered for ease of manufacturing and simple form factor. More complex pad geometries were not considered as there is currently no easy parametric way for us to design and optimize them. 

For the cylindrical silicone gel pads, we utilize three different radii (denoted r). We also consider three different sets of radii (denoted a, b, and c) for the ellipsoid gel pads. Ellipsoid shapes are created in CAD software by non-symmetrically transforming a unit sphere. We then take an appropriately-sized section (width W, length L, and thickness t) from the surfaces of these shapes to create our silicone gel pads.

\begin{figure*}[ht!]
    \centering
    \includegraphics[width=0.9\textwidth]{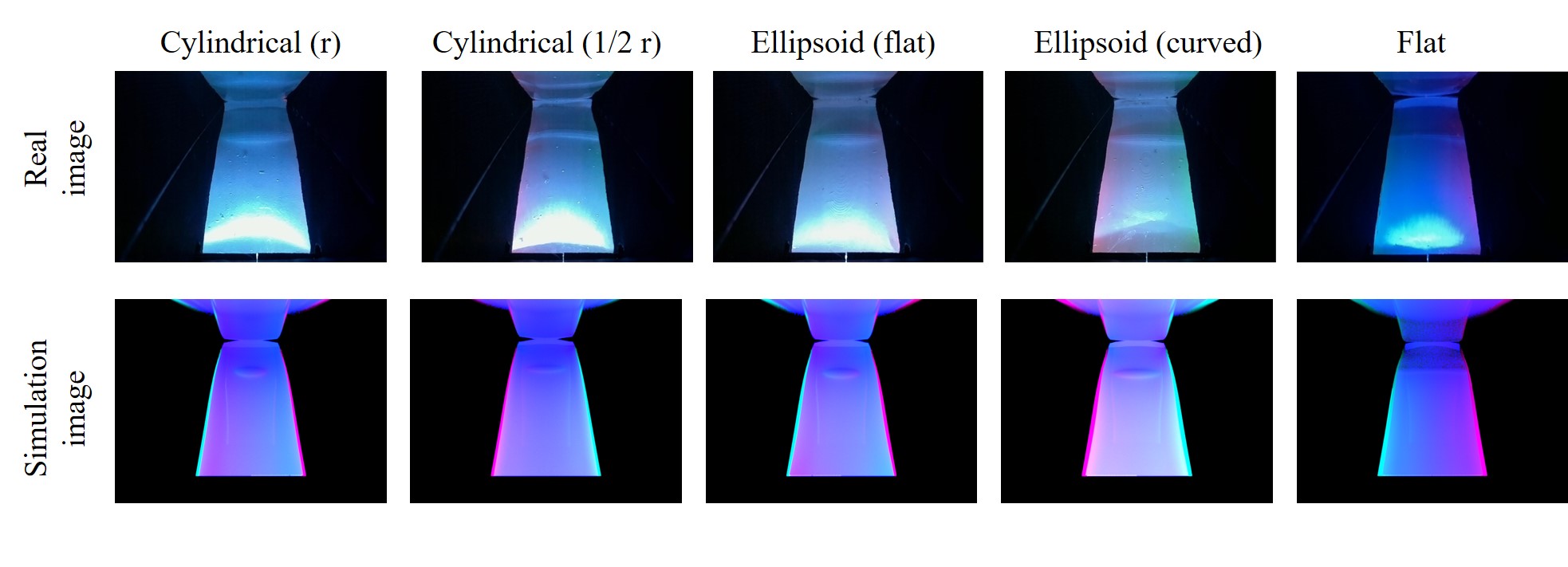}
    \vspace{-15pt}
    \caption{\textbf{Gel Pad variation test}: The top row shows the real-world prototype images captured with \qty{20}{\mm} cylinder indentation. The bottom row shows the simulation results for the same setup using our approach.}
    \vspace{-15pt}    
\end{figure*}

\subsection{Change in the illumination in simulation}
\begin{figure}
    \centering
    \vspace{-3 mm}
    \includegraphics[width=0.9\columnwidth]{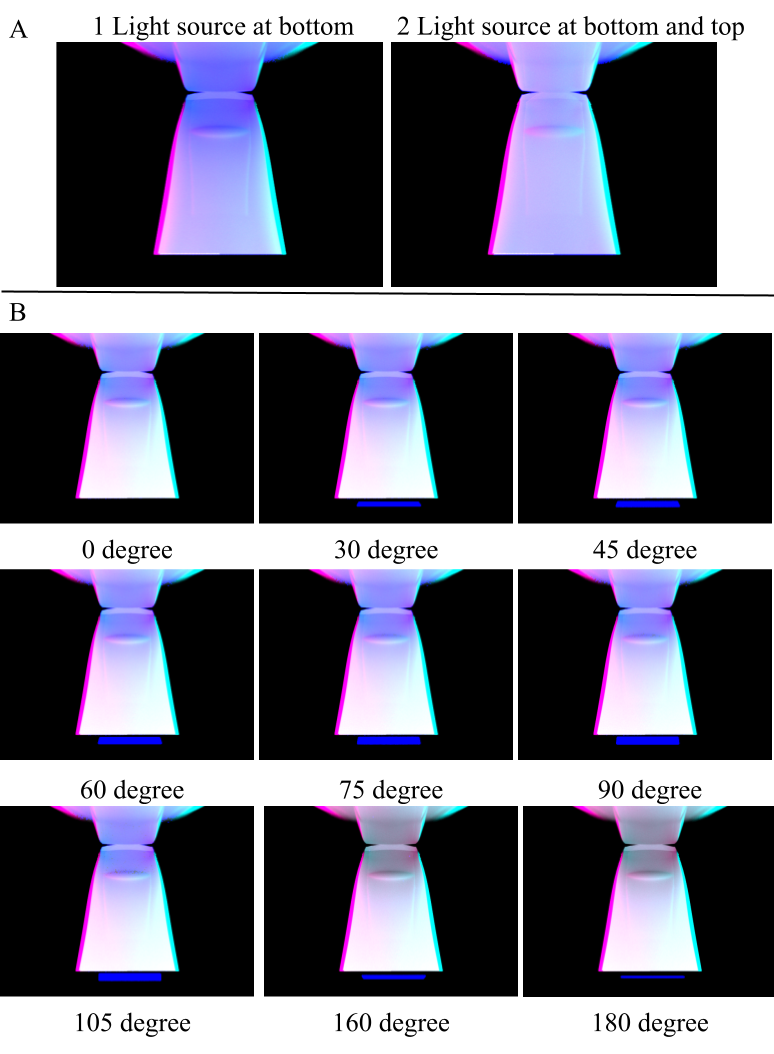}
    \caption{\textbf{Illumination design results}: Part A shows two light conditions -- case 1 when only one blue light is placed at the bottom to illumination the sensor and case 2 when two blue lights are placed, with one at the top and one at the bottom of the sensor to illuminate the gel pad. Part B shows the cases with different incident light panel angles with respect to the gel pad.}
    \label{fig:illum_design}
    \vspace{-5 mm}
\end{figure}

\begin{figure}
    \centering
    \includegraphics[width=\columnwidth]{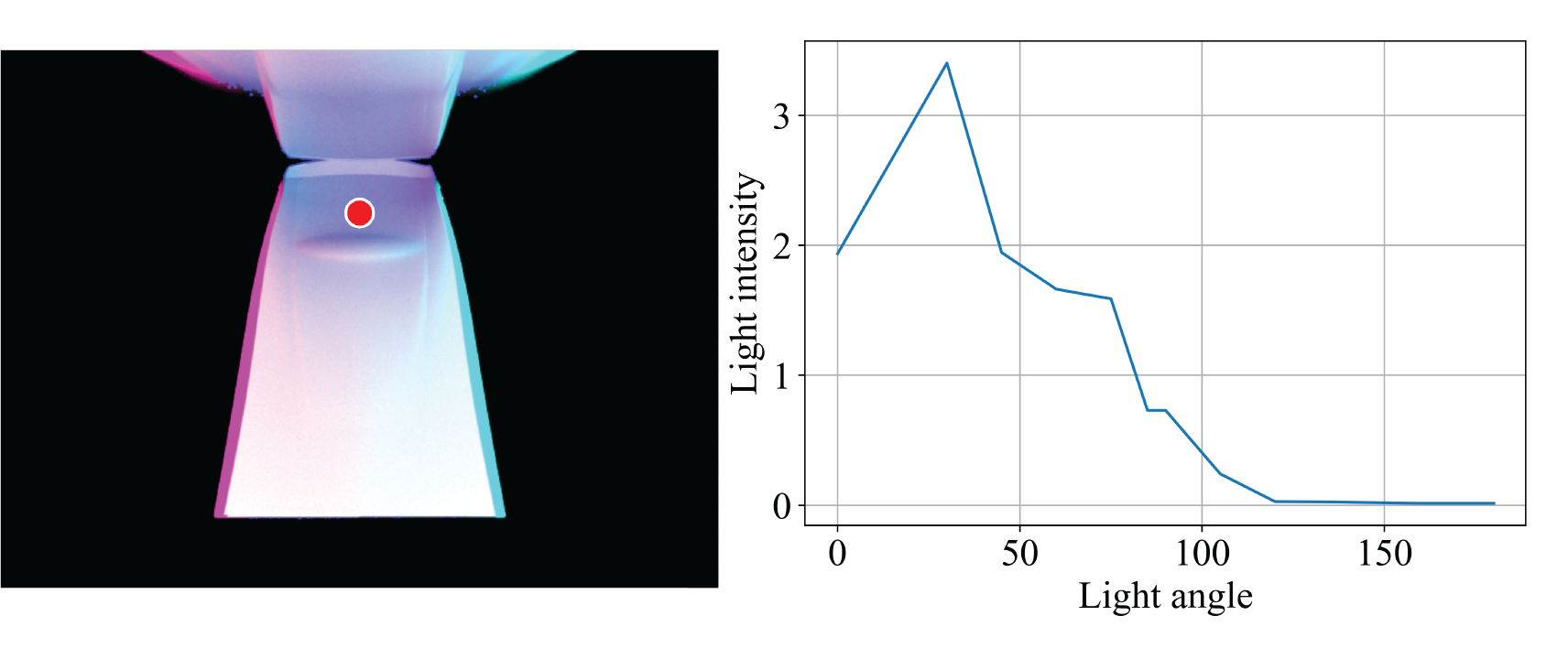}
    \caption{\textbf{Light intensity vs illumination angle}: (Left) shows the point on the indenter which is used for selecting the intensity value in an image. (Right) shows the intensity plot for the red spot for images with  different incident angles of the blue light panel with respect to the gel pad.}
    \label{fig:illum_plot}
    \vspace{-5 mm}
\end{figure}

Choosing where to place the light source and what should be the pose of the light with respect to the gel pad is a trial-and-error process. Therefore, we perform forward design studies using our optical simulation for a given mechanical simulation to obtain the best illumination. Specifically, we tested two variations -- adding another light source at the top and changing the angle of the light source at the bottom to assess the effect on the perception. For the following experiments we chose the flattest ellipsoid gelpad. 
\paragraph{Adding blue at the top} In this experiment, we added another blue light at the top and applied the fluorescent effect appropriately to red and green paint lights. The effect can be seen in Figure \ref{fig:illum_design}A. Having two lights increases the illumination uniformity across the gel pad. As the gel pad shape becomes bigger we think that these additional light sources will be very helpful.  
\paragraph{Changing the light angle} We changed the light source panel incident angle with respect to the gel pad. We plotted the intensity at a point near the indenter as shown in Figure \ref{fig:illum_plot} left. We found that angle=30$^{\circ}$ leads to the highest illumination at the indenter and thus is the best design. Subsequently, the intensity value sharply goes down for angles greater than 90$^{\circ}$.

\subsection{Grasping Tasks}

To show the usefulness of the different simulation-driven Fin Ray designs, we manufacture the different Fin Rays based on the simulation results of the differing gel pads. We end up with a total of three different sizes of Fin Ray tactile sensing pads including the GelSight Baby Fin Ray one. 

In particular, we show that these different Fin Ray shapes can be useful for a variety of different grasping tasks. Smaller Fin Rays, for instance, can reach into narrower spaces and singulate smaller objects. The widest and largest Fin Rays can have a stabler grasp on longer objects, which require higher torque to manipulate. Finally, because all of the Fin Rays are compliant, they can also interact with softer, more fragile objects. 

One phenomenon we noticed was that although the \qty{35}{\mm} $\times$ \qty{70}{\mm} pad visually had the best tri-color sensing region, when we applied a similar gel pad to the widest Fin Ray pad (\qty{50}{\mm} width), the fluorescent colors were not as vivid. However, we were still able to view the tri-color tactile deformation.

\begin{figure}[ht!]
    \vspace{3 mm}
    \centering
    \includegraphics[width=0.87\columnwidth]{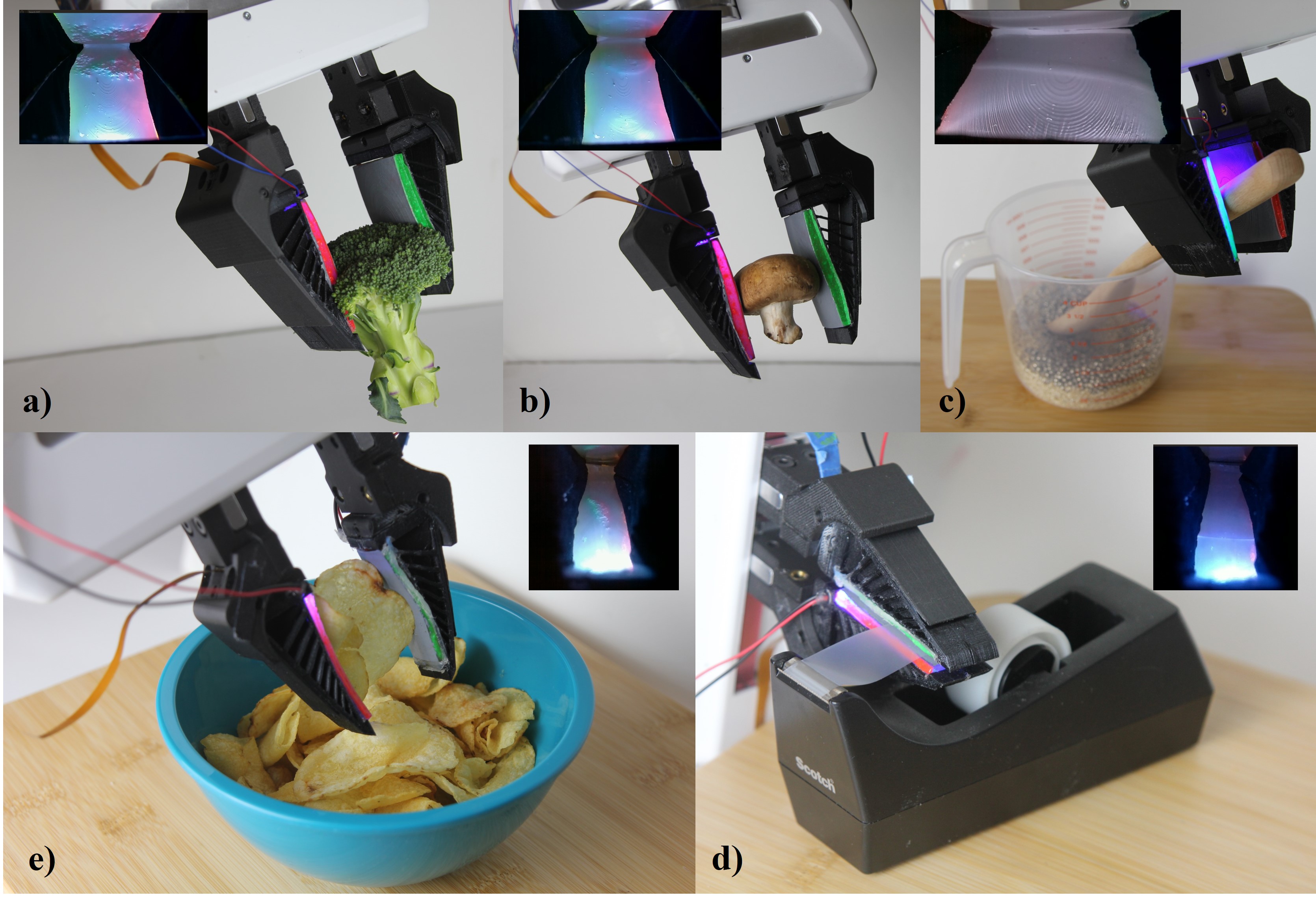}
    \caption{\textbf{Task performing}: The Fin Ray family performing a variety of tasks and their corresponding tactile images. }
    \label{fig:experiments}
    \vspace{-5 mm}
\end{figure}

Overall, we successfully performed simple grasping tasks with all of the different GelSight Fin Rays (Fig. \ref{fig:experiments}). The GelSight Baby Fin Ray was able to grasp a potato chip without crushing it, dig through a bowl of objects to singulate a pecan, and grab tape from a tape dispenser. The widest Fin Ray is able to hold a stirring spoon in a measuring cup full of particulates, and the \qty{35}{\mm} $\times$ \qty{70}{\mm} Fin Ray can grasp a variety of objects such as broccoli, a mushroom, a potato, and a soda can. 

\section{Conclusion}
This paper introduced the first end-to-end simulation framework for compliant GelSight Fin Ray grippers with integrated high-resolution tactile sensing. Through this framework, we are able to explore design spaces much faster than previous methods of prototyping and manufacturing each prototype. We then manufacture our best design with a much wider gel pad that enables grasping a variety of objects - uneven plastic objects, fragile chips, and tiny spatulas. We hope that our work leads to faster design and better adoption of the GelSight Fin Ray for various use-cases.   

\section{Acknowledgment}
This work is financially supported by the Toyota Research Institute and Amazon Science Hub. The authors would also like to thank Yichen Li for his advice in lighting design.
\bibliographystyle{IEEEtran}
\bibliography{IEEEabrv, ref}

\end{document}